\documentclass[runningheads]{llncs}
\usepackage{amsmath}
\usepackage{cite}

\usepackage[T1]{fontenc}
\usepackage{multirow}
\usepackage{graphicx}
\usepackage{booktabs}
\usepackage{makecell}   
\usepackage[caption=false]{subfig}
\begin{document}
\title{CoAL-RAG: A Complexity-Aware Legal Retrieval-Augmented Generation Method}
\titlerunning{CoAL-RAG: Complexity-Aware Legal RAG}
%


\author{Jin Su\inst{1,2} \and
Zhuofeng Zhao\inst{1,2}\thanks{Corresponding author.} \and 
Huanhuan Wang\inst{1,3} \and
Hao Chen\inst{1,3}\thanks{Equal contribution.}}


%
\authorrunning{J. Su et al.}
%
\institute{North China University of Technology, Beijing, 100144, China \and
Beijing Key Laboratory of Key Technologies for AI+ Domain Applications, Beijing, 100144, China \and
Beijing Key Laboratory on Integration and Analysis of Large-scale Stream Data, Beijing 100144, China \\
\email{edzhao@ncut.edu.cn}}

%
\maketitle              
\begin{abstract}
Legal consultation questions exhibit multi-level complexity. A single retrieval strategy often leads to over-reasoning for simple questions and poor interpretability for complex ones, making it difficult to meet the requirements for both answer quality and efficiency in high-risk scenarios. To address this issue, this paper proposes CoAL-RAG, a complexity-aware legal retrieval-augmented generation method, which constructs a multi-dimensional evaluation mechanism based on ``question essence'' and ``retrieval consistency'' to enable adaptive routing of retrieval strategies. First, the reasoning demand is quantified according to the logical structure of the question. Then, the discrepancy between semantic retrieval and keyword retrieval is utilized to indirectly reflect problem complexity, thereby selecting the most appropriate retrieval strategy and dynamically filtering contextual information. Experimental results demonstrate that the proposed method significantly outperforms baseline models not only on Chinese legal benchmarks (SocialLawQA, LawBench) but also demonstrates strong cross-jurisdictional generalization on English datasets (LexGLUE, CaseHold). Specifically, on Chinese datasets, the BLEU score improves by 42.5\% and ROUGE-L reaches 3.6 times that of knowledge graph-based methods. On English benchmarks, CoAL-RAG maintains highly competitive accuracy, achieving an optimal balance between generation quality, deep logical reasoning, and system efficiency across different legal systems.

\keywords{Legal Q\&A \and Retrieval-Augmented Generation \and Complexity Awareness \and Adaptive Retrieval \and Knowledge Graph}
\end{abstract}
\section{Introduction}
Driven by breakthrough advancements of Large Language Models (LLMs) in Natural Language Processing (NLP)~\cite{zhao2023surveyllm,chen2025toolforge}, intelligent legal question answering is transitioning toward a new paradigm of semantic generation. Given the stringent demands for accuracy and traceability in high-stakes legal scenarios~\cite{zhang2025sirens}, integrating external knowledge bases can effectively mitigate model hallucination and knowledge obsolescence~\cite{ji2023hallucination_survey}. However, the complexity of legal consultation queries varies significantly. Questions such as ``What is the statutory retirement age?'' involve a single legal provision and require minimal reasoning, resulting in relatively low complexity. In contrast, queries such as ``My employer fled without signing a labor contract after a workplace injury. Which laws have been violated and how can I protect my rights?'' involve dense legal knowledge, strong conditional constraints, and multi-step reasoning, leading to substantially higher complexity~\cite{duan2019cjrc}.

\begin{figure}[t]
  \centering
  \vspace{-2mm}
  \subfloat[Evidence Gaps in Complex Reasoning Problems\label{fig:ab-a}]{%
    \includegraphics[width=\columnwidth]{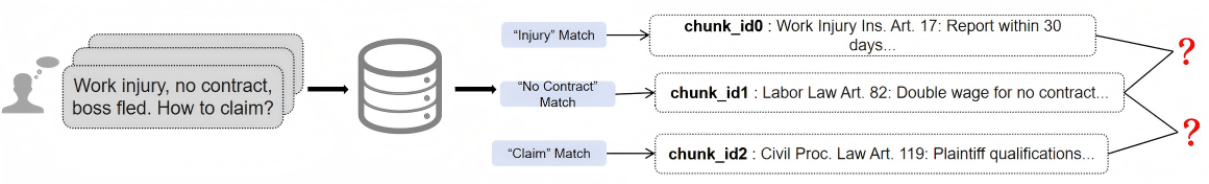}%
  }\\[1mm]
  
  \subfloat[Noise Introduction in Simple Factual Issues\label{fig:ab-b}]{%
    \includegraphics[width=\columnwidth]{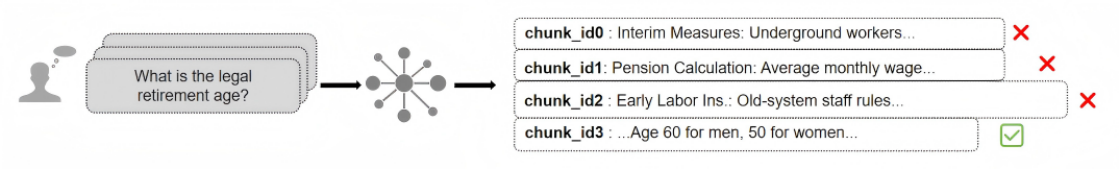}%
  }\\[1mm]
  
  \subfloat[Retrieval conflict\label{fig:ab-c}]{%
    \includegraphics[width=\columnwidth]{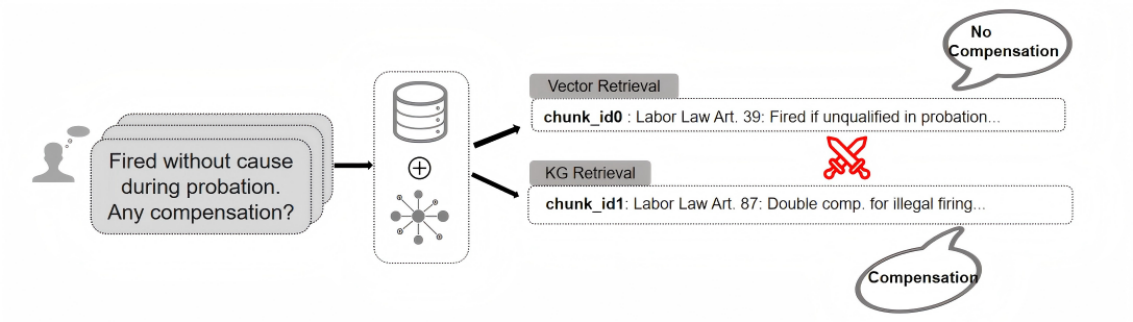}%
  }

  \caption{Limitations of Existing Methods}
  \vspace{-4mm}
  \label{fig:abc}
\end{figure}

Among existing retrieval strategies in the legal domain, pure vector-based retrieval lacks deep relational reasoning, yielding only fragmented legal provisions (Fig.~\ref{fig:ab-a}). Conversely, the indiscriminate application of graph reasoning leads to computational redundancy and noise interference (Fig.~\ref{fig:ab-b}). Furthermore, a direct hybrid of the two often results in inconsistent retrieval outcomes and reasoning conflicts (Fig.~\ref{fig:ab-c}). To address these shortcomings, this paper proposes CoAL-RAG (Complexity-Aware Legal RAG). Leveraging LangGraph, this method constructs a multi-dimensional evaluation mechanism centered on ``question essence'' and ``retrieval consistency''. First, it quantifies the internal logical demands of a query through a five-dimensional metric system. Second, the difference between semantic and BM25 indirectly reflects complexity. Ultimately, it selects the optimal retrieval strategy based on complexity scores and dynamically filters the context, effectively balancing response speed with deep logical accuracy. The main contributions of this paper are summarized as follows:

1) A Multi-dimensional Complexity-Aware Mechanism. We propose a mechanism that evaluates query complexity across multiple dimensions by integrating the internal logic of the query with the external consistency of the retrieval. Furthermore, we design a ``retrieval consistency'' algorithm based on a competition function, providing a criterion for adaptive routing characterized by both numerical stability and probabilistic interpretability.

2) The CoAL-RAG Method. We introduce the CoAL-RAG approach, which synergizes complexity awareness, hybrid retrieval, and knowledge graph coordination. Driven by the multi-dimensional evaluation mechanism, this framework dynamically tailors retrieval strategies to achieve highly efficient and accurate generation for legal question answering.

3) Extensive Cross-Jurisdictional Validation and Performance Trade-off. Experiments conducted on both Chinese civil law datasets (SocialLawQA, LawBench) and English common law benchmarks (LexGLUE, CaseHold) demonstrate that our approach significantly outperforms existing baselines. CoAL-RAG successfully bridges the reasoning gap across different legal jurisdictions, enhancing accuracy for complex queries while maintaining low-latency responses.

\section{Related Works}
\subsection{Legal Large Models}
General-purpose LLMs~\cite{achiam2023gpt4, dubey2024llama3} frequently suffer from legal hallucinations. Early domain models~\cite{chalkidis2020legalbert, xiao2021lawformer, shao2020bertpli} optimized comprehension but lacked generative capabilities. Subsequent instruction-tuned models~\cite{cui2023chatlaw, zhou2024lawgpt, lexilaw2023} and knowledge-augmented models~\cite{huang2023lawyerllama, yue2023disclawllm} improved intent recognition and reasoning. However, they remain inadequate for resolving highly complex, multi-step legal consultations.

\subsection{Retrieval-Augmented Generation}
RAG~\cite{lewis2020rag} mitigates knowledge lag via hybrid retrieval~\cite{m3embedding2024}, re-ranking~\cite{ma2023reranker, fei2024lawbench}, and dynamic routing based on token confidence (FLARE~\cite{jiang2023active_rag}) or generic complexity classifiers (Adaptive-RAG~\cite{jeong2024adaptive_rag}). Despite their success in open-domain tasks, applying these methods to legal queries reveals critical limitations. Token-confidence metrics fail to capture rigorous judicial deduction, and one-dimensional complexity classifiers ignore the multifaceted nature of legal queries. Unlike Adaptive-RAG's black-box routing, CoAL-RAG introduces a transparent, multi-dimensional complexity assessment explicitly tailored to the hierarchical ``chapter-and-clause'' structure of statutory texts, enabling precise dynamic routing aligned with legal reasoning.

\subsection{Knowledge Graph}
Knowledge Graphs (KGs) enhance complex reasoning in RAG~\cite{ji2021kg_survey, wang2023knowledgpt} via path-based explicit chains ( RoG~\cite{luo2023graph_reasoning}, ToG~\cite{sun2023thinkongraph}) or subgraph-based structural extraction~\cite{he2024gretriever, edge2024graphrag, pan2024llm_kg_roadmap}. While effective generally, applying graph structures directly to statutory tasks introduces two main challenges: (1) indiscriminate retrieval introduces noise and latency~\cite{shi2023distracted, trivedi2023interleaving}; and (2) general models like HAKE~\cite{zhang2020hierarchy_kge} fail to capture the hierarchical ``chapter-and-clause'' structure of legal texts, resulting in the loss of fine-grained judicial logic~\cite{zhao2022legal_judgment}.

\begin{figure}
\includegraphics[width=\textwidth]{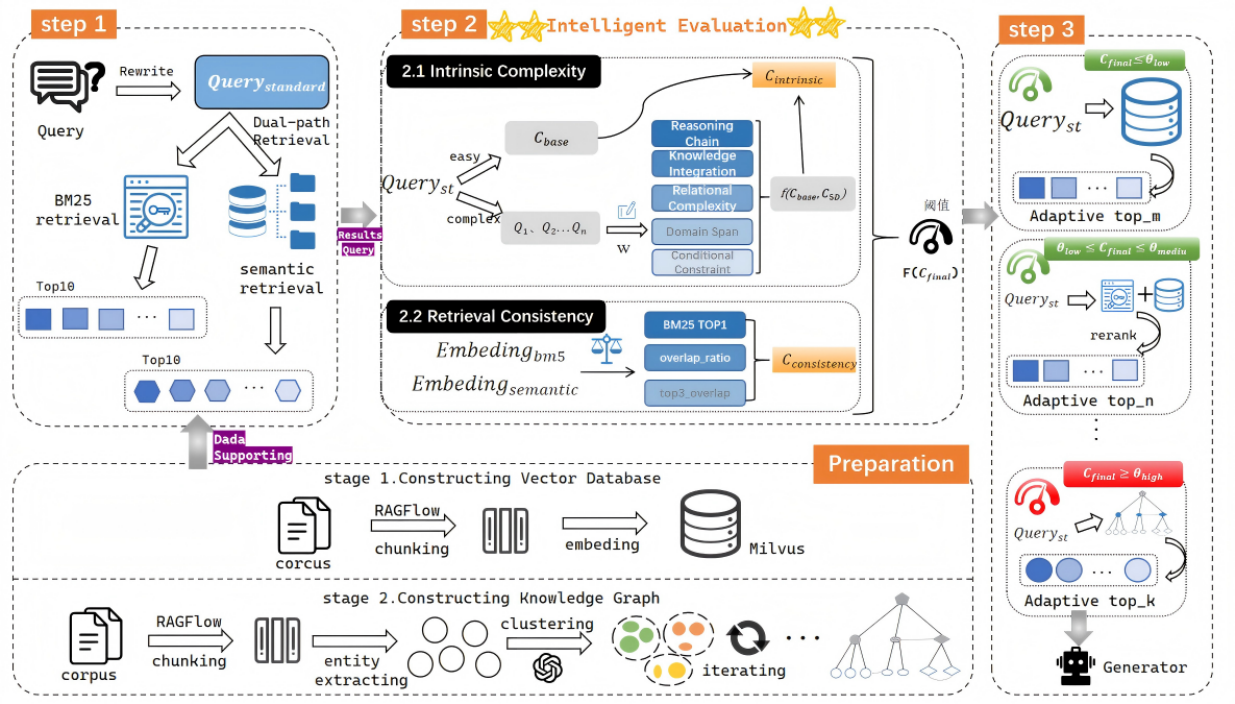}
\caption{Overall Framework of CoAL-RAG} \label{all}
\vspace{-6mm}
\end{figure}

\section{Method}
To address queries with varying levels of complexity, we propose CoAL-RAG. The framework is built upon a hierarchical legal knowledge graph and incorporates complexity-aware modeling together with adaptive context-driven routing to enable dynamic selection of retrieval strategies. The overall system is implemented using LangGraph, as illustrated in Fig.~\ref{all}.

\subsection{Problem Description}
To address the diverse complexity of legal queries, this paper proposes CoAL-RAG, which enables dynamic routing via a complexity-aware mechanism. The input is formally defined as $T=\{Q, D, G\}$, where $Q$ denotes the user's natural language legal query, $D=\{d_1, d_2, \ldots, d_n\}$ represents the unstructured legal corpus, and $G=\{E, R\}$ is a hierarchical knowledge graph in the legal domain. The method computes a complexity score $C_{\text{final}} \in [0,1]$ based on both query characteristics and retrieval consistency. Guided by a threshold $\theta$, it dynamically selects a retrieval strategy $S$, constructs the corresponding context $C_S$, and generates the answer $A = \arg\max_{A'} P(A' \mid Q, C_S)$.

\subsection{Hierarchical Legal KG Construction}
We build a KG $G$ ($\sim$4.2k nodes, $\sim$11.5k edges) across 16 statutes, defining entities (\textit{Law, Chapter, Article, Concept}) and relations (\textit{Subsumption, Reference, Conflict}). The pipeline involves: 1) \textbf{Extraction}: LLMs extract $(e_s, r, e_o)$ triplets from corpus $D$, retaining metadata. 2) \textbf{Fusion}: LLMs resolve contradictions by merging divergent entities into unified nodes. 3) \textbf{Clustering}: BGE-M3 and GMM-UMAP hierarchically group articles (Articles$\rightarrow$Sections$\rightarrow$Domains) to mirror statutory taxonomy. 4) \textbf{Deployment}: Dual-indexed in Milvus and MySQL, the KG supports version-controlled incremental updates.

\subsection{Multi-dimensional Complexity Awareness Mechanism}
This mechanism dynamically integrates existing retrieval methods by evaluating query complexity across multiple dimensions to activate tailored retrieval strategies accordingly. It ensures deep reasoning for complex queries while maximizing response efficiency for simple ones.

\subsubsection{Problem Intrinsic Assessment}
Simple legal queries are assigned a low base complexity through pattern matching, where $C_{\text{base}} \in [0.1, 0.2]$, and their intrinsic complexity is defined as $C_{\text{intrinsic}} = C_{\text{base}}$.

For complex legal queries, semantic features are used to categorize them into six types: ``scenario reasoning'', ``conditional judgment'', ``multi-condition combinations'', ``interest protection'', ``cross-domain'', and ``complex enumeration''. A base complexity $C_{\text{base}}\in[0.4,0.6]$ is predefined for these types, and weight vectors $\boldsymbol{\omega}$ are assigned according to the ``principle of core feature priority'' (e.g., for ``scenario reasoning'', the logical dimension is prioritized). After type determination, the LLM decomposes the query into a set of subqueries $Q_{\text{sub}}$ and evaluates the five complexity dimensions through a structured prompting scheme.\footnote{The prompt guides the LLM to extract subqueries, entities, and constraints to quantify the five dimensions (normalized by $n=4$). }

\textbf{Reasoning Chain Length (RCL), Knowledge Integration Requirement (KIR), and Domain Span (DS)} are each measured relative to a single unit, with complexity exhibiting linear growth as the number of units increases. The calculation formula is defined in Eq.~\ref{eq:score_linear}:

\begin{equation}
\label{eq:score_linear}
\mathrm{Score}_i = \min\!\left(1.0,\; \frac{V_i - 1}{n}\right).
\end{equation}

Where $V_i$ represents the statistical baseline for each dimension:
\begin{itemize}
  \item RCL uses $V_i = \lvert Q_{\text{sub}} \rvert$, reflecting the logical jumps required for the response;
  \item KIR uses $V_i = \max(\lvert C \rvert,\lvert Q_{\text{sub}} \rvert)$, where $C$ is the set of extracted explicit entities, reflecting the density of legal knowledge integration;
  \item DS adopts $V_i = \lvert D_{\text{aug}} \rvert$, representing the size of the identified augmented domain set, reflecting cross-domain integration difficulty.
\end{itemize}
When $V_i \ge n+1$, the saturation value of $1.0$ is reached, and all are judged as complex.

The initial values for \textbf{Relational Reasoning Complexity (RRC)} and \textbf{Conditional Constraint Density (CCD)} are set to $0$, with scores accumulating incrementally as specific logical structures or constraints appear. The calculation formula is defined in Eq.~\ref{eq:score_accum}:

\begin{equation}
\label{eq:score_accum}
\mathrm{Score}_i = \min\!\left(1.0,\; \frac{V_i}{n}\right).
\end{equation}

Where $V_i$ represents the statistical benchmark for each dimension:
\begin{itemize}
  \item RRC takes $V_i = \lvert R_{\text{aug}} \rvert$, the number of unions between explicit logic and implicit scene relationships, reflecting the degree of logical entanglement;
  \item CCD uses $V_i = \lvert C_{\text{const}} \rvert$, representing the number of numerical or temporal constraints, indicating the precision of boundary determination.
\end{itemize}
Both RRC and CCD start at $0$. When $V_i \ge n$, they reach the saturation value of $1.0$ and are both judged as complex.

Based on the multidimensional assessment, the five-dimensional weighted score $C_{\mathrm{5D}}$ is derived from the dimension scores $\mathrm{Score}_i$ and their dynamically assigned weights $\omega_i$, as formulated in Eq.~\ref{eq:c_5d}:
\vspace{-2mm}
\begin{equation}
\label{eq:c_5d}
  C_{\mathrm{5D}}=\sum_{i\in\{\mathrm{RCL,KIR,RRC,DS,CCD}\}} \omega_i \cdot \mathrm{Score}_i
\end{equation}

Combined with the baseline complexity $C_{\mathrm{base}}$, the final intrinsic complexity is calculated via Eq.~\ref{eq:c_intrinsic}:
\begin{equation}
\label{eq:c_intrinsic}
  C_{\mathrm{intrinsic}}=\alpha\, C_{\mathrm{base}} + \beta\, C_{\mathrm{5D}},
\end{equation}
where $\alpha=0.3$ and $\beta=0.7$. The closer $C_{\mathrm{intrinsic}}$ approaches $1.0$, the more intrinsically complex the legal issue becomes, demanding higher reasoning capability.

\subsubsection{Retrieval Consistency Assessment}
Based on the hypothesis that ``simple queries exhibit high consensus across different retrieval viewpoints'', an external feedback mechanism is introduced to detect potential ambiguity and complexity by measuring discrepancies among multiple retrieval pathways. Specifically, BM25 keyword retrieval and vector-based semantic retrieval are performed to obtain candidate document sets $D_{\mathrm{BM25}}$ and $D_{\mathrm{vec}}$, respectively. The retrieval consistency complexity $C_{\mathrm{consistency}}$ is then computed as follows:

\paragraph{Query Simplicity Index (QSI).}
Using the Top-1 score from BM25 as a proxy for literal matching between the query and the knowledge base, we compute the QSI via a sigmoid transformation and treat it as an inverse complexity indicator (Eq.~\ref{eq:qsi}):

\begin{equation}
\label{eq:qsi}
\mathrm{QSI}=\sigma\!\left(\mathrm{Score}^{\mathrm{BM25}}_{\mathrm{top1}}\right)
  =\frac{1}{1+\exp\!\left(-0.5\left(\mathrm{Score}^{\mathrm{BM25}}_{\mathrm{top1}}-12.5\right)\right)}.
\end{equation}
Here, $12.5$ is an empirical threshold. A higher $\mathrm{QSI}$ (approaching $1$) indicates more reliable literal matching and lower retrieval complexity.

\paragraph{Retrieval Divergence Index (RDI).}
This metric quantifies the divergence between the keyword and semantic retrieval result sets by measuring their overlap (Eq.~\ref{eq:rdi}):
\begin{equation}
\label{eq:rdi}
  \mathrm{RDI}=1.0-\left(0.7\,R_{\mathrm{overlap}}\!\left(D_{\mathrm{BM25}},D_{\mathrm{vec}}\right)
  +0.3\,R_{\mathrm{top3}}\!\left(D_{\mathrm{BM25}},D_{\mathrm{vec}}\right)\right).
\end{equation}
Here, $R_{\mathrm{overlap}}(\cdot,\cdot)$ denotes the Jaccard similarity coefficient, and $R_{\mathrm{top3}}(\cdot,\cdot)$ denotes the top-3 document overlap rate. A higher $\mathrm{RDI}$ indicates greater disagreement between semantic understanding and keyword matching.

\paragraph{Consistency Fusion via Competitive Gating.}
To nonlinearly integrate the above metrics, we define the ``simple evidence energy'' $E_{\mathrm{simple}}$ and ``complex evidence energy'' $E_{\mathrm{complex}}$ as follows (Eq.~\ref{eq:energy}):
\begin{equation}
\label{eq:energy}
  E_{\mathrm{simple}}=\mathrm{QSI}^{\,p}\left(1-\mathrm{RDI}\right)^{q}+\varepsilon,\qquad
  E_{\mathrm{complex}}=\left(1-\mathrm{QSI}\right)^{p}\mathrm{RDI}^{\,q}+\varepsilon,
\end{equation}
We set $p=1.5$ to apply a non-linear penalty to low-confidence literal matches (QSI), effectively filtering out weak keyword signals, and $q=0.3$ to maintain a smooth response to retrieval divergence (RDI), preventing minor overlaps from causing routing jitter. The final complexity of the retrieval consistency is computed as the proportion of complex evidence, shown in Eq.~\ref{eq:c_consistency}:

\begin{equation}
\label{eq:c_consistency}
  C_{\mathrm{consistency}}=\frac{E_{\mathrm{complex}}}{E_{\mathrm{complex}}+E_{\mathrm{simple}}}.
\end{equation}

\subsubsection{Unified Complexity Score}
To comprehensively evaluate query complexity, this paper integrates two aspects ``\emph{problem essence}'' and ``\emph{retrieval consistency}''. The final complexity score is formulated in Eq.~\ref{eq:c_final}:

\begin{equation}
\label{eq:c_final}
  C_{\mathrm{final}} = \gamma\, C_{\mathrm{intrinsic}} + (1-\gamma)\, C_{\mathrm{consistency}}.
\end{equation}
We set $\gamma=0.5$ to assign equal importance to the query's linguistic structure and the retrieval system's feedback, ensuring a balanced perspective between internal reasoning demands and external evidence consistency.

\begin{table}[htbp]
\centering
\vspace{-4mm}
\caption{Case Study of the Complexity Awareness in CoAL-RAG}
\label{tab:complexity_case}

\resizebox{\textwidth}{!}{
\renewcommand{\arraystretch}{1.0}
\begin{tabular}{@{}llp{8.5cm}c@{}}
\toprule
\textbf{Phase} & \textbf{Metrics} & \textbf{Explanation} & \textbf{Final Score} \\
\midrule

\multirow{6}{*}{\makecell[l]{\textbf{$C_{\text{intrinsic}}$}}}

& \textbf{Base Setup} ($C_{\text{base}}=0.50$) 
& Type: Scenario Reasoning,a unified weight $\omega=0.25$. 
& \multirow{6}{*}{\textbf{0.50}} \\
& \textbf{RCL} (0.75)
& 4 sub-queries (invention/relevance/ownership/time) & \\
& \textbf{KIR} (1.00)
& 8 entities (Zhang San/A Company/PC/patent, etc.) & \\
& \textbf{DS} (0.50)
& 3 domains (Patent / Labor / Civil Code) & \\
& \textbf{RRC} (0.75)
& 3 relations (employment / ownership / infringement)& \\
& \textbf{CCD} (1.00)
& 4 constraints (weekend/company PC/non-core/resigned)& \\
\midrule
\multirow{2}{*}{\makecell[l]{\textbf{$C_{\text{consistency}}$}}}
& \textbf{QSI} (0.32)
& $Score_{BM25}^{top1}=11.0$(low literal match degree)
& \multirow{2}{*}{\textbf{0.871}} \\

& \textbf{RDI} (0.93)
& $R_{top3}=0.0$, $R_{overlap}=0.1$(significant divergence between semantic and BM25)& \\

\bottomrule
\end{tabular}
}
\vspace{-6mm}
\end{table}

\subsection{Dynamic Retrieval Routing and Adaptive Context Construction}
To accommodate queries of varying complexity, this paper introduces a complexity-aware multi-path routing strategy. Based on three predefined thresholds($\theta_{\text{low}}=0.25$, $\theta_{\text{medium}}=0.45$ and $\theta_{\text{high}}=0.7$), the processing pipeline is organized into four distinct tiers:

When $C_{\text{final}} \le \theta_{\text{low}}$, the query is classified as a simple factual question. Dense vector retrieval is activated, relying on the large model's intrinsic reasoning capabilities to generate answers efficiently.

When $\theta_{\text{low}} < C_{\text{final}} \le \theta_{\text{medium}}$, the query is considered semantically ambiguous and requiring precise localization. A hybrid retrieval strategy combining dense and sparse methods is employed, followed by a re-ranking module to refine results and mitigate semantic drift. Iterative processing is also applied to enhance answer accuracy.

When $\theta_{\text{medium}} < C_{\text{final}} \le \theta_{\text{high}}$, the query is treated as moderately complex and handled via network graph retrieval.

When $C_{\text{final}} \ge \theta_{\text{high}}$, the query is identified as highly complex, graph--text verification is activated. Using the logical reasoning paths extracted from the hierarchical knowledge graph as the backbone, the legal provision fragments retrieved through hybrid retrieval are cross-validated to eliminate conflicting texts.

After determining the retrieval strategy, adaptive truncation based on score cliffs is applied to further reduce tail noise. The score decline rate between adjacent documents is defined as $\Delta_i = (s_i - s_{i+1}) / s_i$, with a cliff threshold $\sigma = 0.2$. The optimal truncation position is determined as the first index where the decline exceeds $20\%$, so $k = \arg\min_i \{\Delta_i > \sigma\}$. The resulting context set $C_{\text{ctx}} = \{d_1, d_2, \ldots, d_k\}$ is then incorporated into the prompt template to guide the final answer generation.

\section{Experimental}
The domain of social law encompasses high-frequency scenarios such as labor contracts and work injury identification. Its well-defined structure and hierarchy make it ideal for evaluating the adaptability of CoAL-RAG. Experiments are conducted on the self-constructed SocialLawQA dataset and the public LawBench benchmark.

\noindent\textbf{Hyperparameter Calibration} Hyperparameters ($\alpha, \beta, p, q,\gamma$) and thresholds ($\theta$) were calibrated via grid search on an expert-annotated, stratified validation set ($N=120$). Sensitivity analysis (Sec. 5.4) shows performance remains stable within $\pm 10\%$ parameter variance, confirming the robustness of our complexity-aware design.

\subsection{Baselines}
To evaluate the effectiveness of CoAL-RAG, we compare it with the following baselines: 
1) \textbf{Inference without Retrieval}: Direct inference, Chain-of-Thought (CoT) reasoning~\cite{wei2022chain} and LawGPT\_zh~\cite{zhou2024lawgpt}.
2) \textbf{Inference with Retrieval}: Retrieval-Augmented Generation (RAG)~\cite{lewis2020rag}, Hybrid RAG~\cite{guu2020realm}, CLERAG, IRCoT~\cite{trivedi2023interleaving}, and Search-o1~\cite{li2025search}. 
3) \textbf{Unified Retrieval-and-Reranking Models}: bge-reranker-v2-m3 and Qwen3-Reranker-4B. 
4) \textbf{Knowledge Graph Augmentation}: G-Retriever~\cite{he2024gretriever} (flat graph) and LeanRAG~\cite{zhang2025leanrag} (hierarchical graph). 
5) \textbf{RL Tuning Methods}: R1~\cite{guo2025deepseek} and Search-R1~\cite{jin2025search}. R1 performs reasoning based on internal knowledge, while Search-R1 interacts with a search engine during inference. 
For fairness, all RL methods use the F1 score as the reward metric and follow their original training settings. Real-world retrieval is simulated using Google Web Search via SerpAPI, with ten retrieved documents for each method.




\subsection{Evaluation Metrics}
We assess generation quality on Chinese legal benchmarks using \textbf{ROUGE (1/2/L)}, \textbf{BLEU-4}, and \textbf{BERTScore} to measure token overlap and deep semantic alignment. For English cross-jurisdictional benchmarks, we report \textbf{Accuracy} (for CaseHold) alongside \textbf{Micro-F1} and \textbf{Macro-F1} (for LexGLUE) to evaluate multi-class logical reasoning performance. Finally, system efficiency is measured via Average Response Time (ART), with detailed latency analysis presented in Section 5.2.

\subsection{Datasets}
Chinese Benchmarks (Civil Law): \textbf{SocialLawQA} is a curated dataset of 1.5k Q\&A pairs across 16 statutes, featuring a diverse complexity distribution ideal for validating adaptive routing in real-world scenarios. \textbf{LawBench}~\cite{jiang2023active_rag} is an authoritative benchmark from which we selected a 1k Q\&A subset focusing on social law to evaluate core dimensions like memory, comprehension, and application.

English Benchmarks (Common Law): To evaluate cross-jurisdictional adaptability, we utilize \textbf{LexGLUE}~\cite{chalkidis2022lexglue} (specifically subsets requiring logical deduction) for broad legal NLU assessment, and \textbf{CaseHold}~\cite{Zheng2021casehold}, a challenging multiple-choice dataset rigorously testing long-text reasoning and legal holding identification.


\section{Results}
\subsection{Generation Quality and Generalization}
Table~\ref{tab:main_results} compares CoAL-RAG with baselines (utilizing Qwen2.5-3B-Instruct as the primary base model unless otherwise specified).

\textbf{Chinese Benchmarks (Civil Law):} Pure parametric models perform poorly due to domain hallucinations. Static pipelines suffer from semantic drift, while pure KG methods (LeanRAG) introduce noise. In contrast, CoAL-RAG achieves the highest BLEU scores (0.2815 on LawBench, 0.1684 on SocialLawQA), delivering highly competitive accuracy comparable to the compute-heavy Search-R1, with improvements in key precision metrics being statistically significant ($p < 0.05$).

\begin{table}[t]
\caption{Comprehensive Results. R-1/2/L: Rouge-1/2/L; BL: BLEU; BS: BERTScore; Mi-F: Micro-F1; Ma-F: Macro-F1; Acc: Accuracy. $^\star$Out-of-domain. '-' indicates system/language mismatch or excessive migration cost. \textbf{Bold} and \underline{underlined} are best results. $\dagger$ denotes statistical significance ($p < 0.05$) over the strongest baseline via paired t-test.}
\label{tab:main_results}
\centering
\footnotesize 
\setlength{\tabcolsep}{2.2pt}
\renewcommand{\arraystretch}{1.05} 
\resizebox{\textwidth}{!}{%
\begin{tabular}{l ccccc ccccc cc c}
\toprule
\multirow{3}{*}{\textbf{Methods}} & \multicolumn{10}{c}{\textbf{Chinese Benchmarks (Civil Law)}} & \multicolumn{3}{c}{\textbf{English Benchmarks (Common Law)}} \\
\cmidrule(lr){2-11} \cmidrule(lr){12-14}
 & \multicolumn{5}{c}{LawBench$^\star$} & \multicolumn{5}{c}{SocialLawQA$^\star$} & \multicolumn{2}{c}{LexGLUE} & CaseHold \\
\cmidrule(lr){2-6} \cmidrule(lr){7-11} \cmidrule(lr){12-13} \cmidrule(lr){14-14}
 & R-1 & R-2 & R-L & BL & BS & R-1 & R-2 & R-L & BL & BS & Mi-F & Ma-F & Acc \\
\midrule
\multicolumn{14}{l}{\textit{\textbf{Qwen2.5-3B-Instruct}}}\\
Direct Inference & 0.2523 & 0.0718 & 0.1670 & 0.0254 & 0.7220 & 0.2003 & 0.0426 & 0.1132 & 0.0298 & 0.7351 & 0.2810 & 0.2220 & 0.4954 \\
CoT         & 0.2680 & 0.0543 & 0.1673 & 0.0312 & 0.7524 & 0.3369 & 0.1266 & 0.2159 & 0.0628 & 0.7765 & 0.3056 & 0.2452 & 0.5126 \\
LawGPT\_zh  & 0.2677 & 0.0691 & 0.2046 & 0.0294 & 0.7548 & 0.3461 & 0.1273 & 0.2186 & 0.0638 & 0.7866 & - & - & - \\
\midrule
Standard RAG    & 0.4022 & 0.2456 & 0.3257 & 0.1739 & 0.8002 & 0.3781 & 0.1855 & 0.2763 & 0.1240 & 0.7972 & 0.4550 & 0.4937 & 0.5211 \\
Hybrid RAG  & 0.4212 & 0.2534 & 0.3339 & 0.1975 & 0.8058 & 0.3799 & 0.1777 & 0.2833 & 0.1059 & 0.7965 & 0.6782 & 0.6150 & 0.5320 \\
CLERAG      & 0.4258 & 0.2561 & 0.3130 & 0.1588 & 0.8140 & 0.4003 & 0.2104 & 0.2692 & 0.1073 & 0.8065 & 0.6617 & 0.6020 & 0.6251 \\
IRCoT       & 0.3244 & 0.1021 & 0.2055 & 0.0532 & 0.7800 & 0.3936 & 0.1978 & 0.2533 & 0.1015 & 0.7843 & 0.3858 & 0.3942 & 0.5102 \\
Search-o1   & 0.2666 & 0.0772 & 0.1748 & 0.0497 & 0.7611 & 0.3345 & 0.1574 & 0.2688 & 0.1014 & 0.8005 & 0.3142 & 0.2683 & 0.5037 \\
\midrule
bge-rerank  & 0.4477 & 0.3166 & 0.3679 & 0.2424 & 0.8258 & 0.4299 & 0.2621 & 0.3210 & 0.1355 & 0.8301 & 0.5936 & 0.4339 & 0.6550 \\
Qwen3-Rerank& 0.3031 & 0.1284 & 0.2145 & 0.0913 & 0.7756 & 0.3644 & 0.1634 & 0.2622 & 0.0939 & 0.8010 & 0.3712 & 0.5195 & 0.5383 \\
\midrule
G-retriever & 0.2450 & 0.0687 & 0.1753 & 0.0466 & 0.7480 & 0.3220 & 0.1258 & 0.2267 & 0.0650 & 0.7845 & 0.2603 & 0.4047 & 0.6232 \\
LeanRAG     & 0.1911 & 0.0466 & 0.1137 & 0.0188 & 0.7327 & 0.2315 & 0.0629 & 0.1204 & 0.0237 & 0.7480 & - & - & - \\
\midrule
Search-R1   & \underline{\textbf{0.4953}} & 0.3527 & \underline{\textbf{0.4430}} & 0.2700 & \underline{\textbf{0.8419}} & 0.4320 & \underline{\textbf{0.2738}} & \underline{\textbf{0.3321}} & 0.1558 & \underline{\textbf{0.8404}} & 0.6925 & \underline{\textbf{0.6585}} & 0.6745 \\
\midrule
\textbf{CoAL-RAG (Ours)}& 0.4832 & \underline{\textbf{0.3690}}$^{\dagger}$ & 0.4177 & \underline{\textbf{0.2815}}$^{\dagger}$ & 0.8342 & \underline{\textbf{0.4427}}$^{\dagger}$ & 0.2560 & 0.3302 & \underline{\textbf{0.1684}}$^{\dagger}$ & 0.8184 & \underline{\textbf{0.7186}}$^{\dagger}$ & 0.6520 & \underline{\textbf{0.6885}}$^{\dagger}$ \\
\bottomrule
\end{tabular}}
\end{table}

\textbf{English Benchmarks (Common Law):} Evaluated on LexGLUE and CaseHold, pure parametric models predictably struggle without common-law grounding (e.g., 0.4954 CaseHold Accuracy). Conversely, CoAL-RAG exhibits robust generalization, outperforming the RL-tuned Search-R1 (0.6885 Accuracy, 0.7186 Micro-F1). Despite a marginal Macro-F1 lag due to our Civil-Law-centric KG lacking precedent indexing, CoAL-RAG consistently surpasses generic rerankers and Hybrid RAG without requiring costly reinforcement learning.

\subsection{Efficiency Analysis}
Benefiting from complexity-aware routing, CoAL-RAG avoids redundant computation for simple queries, achieving average response times of 4.76\,s and 5.09\,s on LawBench and SocialLawQA. It is $\sim$2.2$\times$ faster than LawGPT and faster than complex graph methods like LeanRAG. While adding $\sim$2\,s of latency compared to Standard RAG, it improves LawBench BLEU by 61.8\%, achieving an optimal trade-off between generation quality and real-time system performance.

\subsection{Ablation Study}
To validate the core components of CoAL-RAG, we conduct ablation experiments on LawBench using three variants: (1) w/o Intrinsic---removes intrinsic complexity assessment, relying solely on retrieval consistency for routing; (2) w/o Consistency---omits retrieval consistency assessment, using only query features; and (3) w/o Dynamic---replaces adaptive document selection with a fixed Top-10 set for generation. To rule out random variance, all scores are averaged across multiple runs.

Two additional metrics, Article F1 and LawConcept Recall (measuring statutory article retrieval and legal concept coverage), are introduced in the ablation study. Table~\ref{tab:ablation_results} presents the ablation results.
\subsubsection{The Effectiveness of Intrinsic Assessment}
The results show that removing intrinsic complexity assessment leads to a decline in Article F1 from 0.5308 to 0.4977 (a relative decrease of 6.24\%), underscoring the importance of anticipating logical depth for identifying complex queries. ROUGE-L also decreases from 0.4162 to 0.4025, indicating that fine-grained perception of question types contributes to the structural coherence and relevance of generated answers. Overall, intrinsic complexity assessment plays a key role in accurately determining query complexity and ensuring effective legal provision retrieval.
\subsubsection{The Effectiveness of Retrieval Consistency}
Removing retrieval consistency evaluation leads to performance declines across all metrics: Article F1 drops by 1.48 percentage points, ROUGE-L by 1.55 points, and both LawConcept Recall and BERTScore also decrease. These results confirm that this module effectively filters retrieval noise and improves routing accuracy through multi-perspective consistency.

\begin{table}[t]
\caption{Ablation Experiment Results on the LawBench Dataset}\label{tab:ablation_results}
\centering
\small
\setlength{\tabcolsep}{4pt} 
\renewcommand{\arraystretch}{1.1}

\resizebox{\textwidth}{!}{%
\begin{tabular}{c|c|c|c|c}
\hline
Methods & Article F1 & \makecell{LawConcept \\ Recall} & ROUGE-L & BERTScore \\
\hline
CoAL-RAG(Ours)   & \underline{\textbf{0.5308}} & \underline{\textbf{0.8146}} & \underline{\textbf{0.4162}} & \underline{\textbf{0.7706}} \\
w/o Intrinsic    & 0.4977 & 0.8059 & 0.4025 & 0.7688 \\
w/o Consistency  & 0.5160 & 0.8071 & 0.4007 & 0.7676 \\
w/o Dynamic      & 0.5278 & 0.8034 & 0.4029 & 0.7653 \\
\hline
\end{tabular}%
}
\vspace{-4mm}
\end{table}

\subsubsection{The Effectiveness of Dynamic Top-K}
The w/o Dynamic variant, despite retrieving a fixed top-10 documents, achieves the lowest LawConcept Recall (0.8034) and BERTScore (0.7653), confirming that indiscriminately increasing document volume introduces noise. In contrast, CoAL-RAG dynamically filters low-relevance documents, improving semantic accuracy while preserving information density.

Ablation results confirm the synergistic effect of the three core modules: multi-dimensional evaluation combined with dynamic filtering enables precise identification of key evidence, effectively balancing answer quality and response efficiency.

\subsection{Sensitivity Analysis}
Sensitivity analysis on a stratified $N=120$ validation set confirms the model's robustness: shifting routing thresholds $\theta_{\{low, med, high\}}$ or the fusion weight $\gamma$ by $\pm 10\%$ causes minimal ($<1.5\%$) fluctuation in ROUGE-L and BERTScore. Gating exponents $p, q$ also exhibit high stability, with performance variance $<1.0\%$ across tested ranges ($p \in [1.2, 1.8], q \in [0.2, 0.4]$). Furthermore, Average Response Time remains consistent ($\pm 0.4s$) even under critical threshold variations, proving that CoAL-RAG's efficacy stems from its structural complexity-aware logic rather than heuristic hyperparameter over-tuning.

\subsection{Case Study and Error Analysis}
Table~\ref{tab:case_study} demonstrates CoAL-RAG's effectiveness across complexities. Error analysis reveals two primary failure modes: (1) \textbf{Knowledge Gaps}: Queries involving local policies (Case 3) outside the statutory KG cause reasoning voids. (2) \textbf{Priority Conflicts}: The model occasionally struggles to resolve hierarchical logic among overlapping laws (e.g., General vs. Special laws). This indicates that the routing mechanism may systematically misclassify edge cases where implicit legal hierarchy is required but not explicitly encoded in the KG. Future work should focus on integrating multi-tier policy data and enhancing legal hierarchy awareness.

\begin{table*}[t]
\centering
\footnotesize
\caption{Representative case analysis of CoAL-RAG under different query complexities.}
\label{tab:case_study}

\resizebox{\textwidth}{!}{
\begin{tabular}{p{5.2cm} >{\centering\arraybackslash}p{2.5cm} p{8.3cm}}
\toprule
\textbf{Query} & \textbf{$C_{\text{final}}$} & \textbf{CoAL-RAG} \\
\midrule

Can I resign during probation without violating the labor contract? 
& $0.31$ 
& Yes. Under the Labor Contract Law, an employee may terminate the contract during the probation period by giving advance notice (typically three days) to the employer (\textbf{Success. Hybrid Retrieval}). \\

Who owns the patent if software is developed after work hours using company equipment?
& $0.79$
& If the invention is not related to the employer's business scope and is not part of assigned duties, the patent rights generally belong to the individual developer rather than the employer (\textbf{Success. Graph Reasoning}). \\

How should eligibility for government housing benefits be determined?
& $0.26$
& The system lacks local policy data and fails to resolve priorities among overlapping administrative regulations. (\textbf{Failure} Knowledge Gap). \\

\bottomrule
\end{tabular}
}
\vspace{-6mm}
\end{table*}

\section{Conclusion}
This paper proposes CoAL-RAG, a multi-dimensional complexity-aware retrieval-augmented generation method tailored to the variable complexity of legal queries. By jointly assessing reasoning depth across multiple dimensions and incorporating retrieval consistency, the approach dynamically selects optimal retrieval strategies and adaptively constructs context. Evaluations across Chinese (Civil Law) and English (Common Law) benchmarks confirm that this complexity-aware mechanism is highly generalizable, mitigating the signal-to-noise trade-off in complex cross-jurisdictional scenarios.

While CoAL-RAG balances quality and efficiency, challenges remain in dynamic adaptation. Future work will: (1) extend to specialized domains (e.g., criminal law, finance); (2) enhance cross-document reasoning to resolve conflicts among overlapping provisions; and (3) scale to larger LLMs (e.g., 7B/14B) to investigate performance ceilings.
%
%
\bibliographystyle{splncs04}
\bibliography{paper}

\end{document}